# Origami Inspired Solar Panel Design

Chris Whitmire
Institute of Artificial Intelligence
chris.whitmire@uga.edu

Brij Rokad
Institute of Artificial Intelligence
brijrokad@uga.edu

Caleb Crumley
Department of Computer Science
crumleyc@uga.edu

**Abstract -** The goal of this paper was to take a flat solar panel and make cuts on the panel to make smaller, but still viable solar panels. These smaller solar panels could then be arranged in a tree-like design. The hope was that by having solar panels faced in different directions in 3-dimensional space, the tree system would be able to pick up more sunlight than a flat solar panel. The results were promising, but this project did not take every factor into account. Specifically, optimum shape, temperature and the system's resistance, reflection of sun-rays were not explored in this project. This project will take an approach from origami paper folding to create the optimum arrangement that will allow the overall system to absorb the maximum energy. Since the system stays stationary throughout the day, it can reduce the maintenance cost and excess energy use because it does not require solar tracking. This project will implement a variety of Evolutionary Algorithms to find the most efficient way to cut a flat solar panel and arrange the resulting smaller panels. Each solution in the population will be tested by computing the amount of solar energy that is absorbed at particular times of the day. The EA will be exploring different combinations of angles and heights of the smaller panels on the tree such that the system can produce the maximum amount of power throughout the day. Our Evolutionary algorithms' performance are comparable to the performance of flat solar panels.

## I. Introduction

Solar energy is one of the most reliable and renewable energy sources. People have been harnessing the sun's power since the 7th century B.C. It's been a millennium and there is still enormous potential left for solar energy to grow. In 1839 scientist Edmond Becquerel started experimenting on electrolytic cells and discovered the photovoltaic effect. In 1908 the first solar collector was produced based on paper of photoelectric effect [6] by Albert Einstein in which he won the Nobel Prize. From that point on, numerous scientists and engineers have worked on solar energy with the goal of making it more efficient and accessible to everyone. Today, solar panels have developed to the extent that a collection of them can self-sustain a house, but there is still much room left for its growth. These days, most of the solar panels are made as a flat panel which stays stationary. Because of this, they are exposed to at most 4 - 5 hours of usable sunlight in a day. Scientists and engineers have developed a sun tracking system, which tracks the sun in real-time and moves the solar panel accordingly with the help of the motor control system. This gets more hours of usable sunlight, but the relative energy produced (Energy produced by solar panel - Energy consumed by motor control system) could be improved. To overcome this problem, we have developed a tree structure solar panel design. Our idea is inspired from origami art and origami inspired solar panels [7]. Our system takes a flat solar panel and makes cuts on it. These sub - solar plates are arranged with different height, axial angle, and orientation such that they get the maximum amount of usable sunlight while being stationary. This way more solar energy can be produced without using the motor control system or using more solar panels. One paper that was critical in inspiring the tree-shaped design[1] has used ice-cracking method which helped us to begin our research. In this paper, we have applied different kinds of evolutionary algorithms to see which strategy is more appropriate for solar panel design. The next section of this paper will dive into the relevant works section citing the research that inspired this experiment. The following sections will explain our methods with the different evolutionary algorithms and the specific parameters for each and their corresponding results. Lastly, this paper will finish up with a summary of this experiment and any suggested future work to better further this research.

## II. Literature Survey

It would be very beneficial, if a more compact and efficient solar panel design could be made. The authors in [1] contributed to this work. Origami is known for their creative design. Authors have tried to establish a structure which can be foldable and deployable by using genetic algorithms. The paper focuses on automatically deriving the crease pattern to get a foldable shape with the help of a genetic algorithm that aims to develop origami structures featuring optimal geometric properties. To determine crease patterns, authors have used the "ice-cracking" method in which crease will form like cracks extending to form forks in ice. Afterwards, a genetic algorithm is applied to encode geometric shape. While in [2], the authors have used a new measure of creativity as a guide in an interactive evolutionary art task, and they tie the results to natural language usage of the term "creative". The generative ecosystemic art system is used, which is called EvoEco- an agent-based pixel-level means of generating images. The best things about this is that the authors have developed an artistic engine which is capable of autonomously generating a wide array of novel images and evolving them via an Interactive Evolutionary Algorithm. This IEA, with the help of user study, evaluates the system comparing the augmented versions, and judging the success of their approach.

The most widely used renewable energy sources are solar panels and wind generators. In [3] authors have design a hybrid system of photovoltaic and wind generators (PV/WG) using genetic algorithms to get an optimal power output. One of the problems were the different power sources. Photovoltaic system produces DC power whereas Wind Generator produces AC power. To overcome this, they have designed a third system which converts the respective power source and stores them in a suitable battery bank. A GA was used to minimize the cost of the entire system.

Origami is an ancient art with a complex structure and design, but it is nonetheless a creative art. In [4] authors have tried to implement the art of origami in active material. The active material does not need an external force to move, it can fold and unfold by itself. With active material, a suitable geometry allows engineering to create self-folding structure which can be used in space systems, underwater robotics, small scale devices, self-assembling systems, and designing dynamic solar system, among other things.

It was not until the last fifteen or so years that it was possible to predict the performance of photovoltaic systems. The authors in [5] developed a method to validate and calibrate the two popular performance models of photovoltaic systems, Sandia Photovoltaic Array Performance Model (SAPM) and the California Energy Commission (CEC) model. SAPM comprises a set of expressions for the short-circuit current, open circuit voltage, and maximum power point. The CEC models a module as a single diode equivalent circuit. The development of these models have cut down on the cost of having to monitor a system outside or with a simulation to conduct any information about the power produced. They have made a huge impact on the cost of photovoltaic research, specifically making this project possible. These models are the two performance models used in our fitness function for our evolutionary algorithms. The authors of [5] were able to cut the error rate in half when calibrating these models with measured temperature coefficients instead of the traditional method at the time. Illustrating the validity and accuracy of these models are important to not only justify our usage, but also to show how far the field itself has come in terms of improving green energy.

In order to get the maximum solar irradiance, the author of the article [8] has tried to get an optimal surface tilt angle. To do that the solar panel is faced south the equator, and the surface angle value is measured by tacking daily global and diffuse solar radiation on a horizontal surface. By doing this, the author got the optimal surface tilt angle for each month. The optimal angle found for August was 12 degrees which is similar to our pvlib fitness function. As shown in table 1 and figure 1, our pvlib fitness got the energy output of 652.47 W while facing south at 15 degrees. Which makes our pvlib fitness function reliable and trustworthy.

TABLE 1. Flat Panel Energy output

| Orientation (N.E.S.W) | Fitness at Surface Tilt (0,15, 30, 45, 60) | | | | |
|---|---|---|---|---|---|
| | 0° | 15° | 30° | 45° | 60° |
| North | 666.40 | 650.67 | 597.94 | 505.73 | 366.38 |
| East | 666.40 | 720.69 | 728.71 | 691.72 | 609.29 |
| South | 666.40 | 652.47 | 612.35 | 541.70 | 442.27 |
| West | 666.40 | 577.56 | 482.52 | 401.17 | 340.29 |

### III. Methodology

We have applied three different evolutionary algorithms to compare the results of each one. Our representation has a 16 bit of array which represents where a cut could be made on a typical solar panel of size 65 x 39 inches. The size of every photovoltaic cell is 6 x 6 inch, thus length-wise there are 10 photovoltaic cells and width-wise there are 6 photovoltaic cells, which makes the total of 60 photovoltaic cells in a solar panel. The first 10 bits of our representation are dedicated to length and last 6 bits are dedicated to width, so every bit represents an array of PV cell lengthwise and widthwise respectively. More specifically, the first bit in our representation represents a possible cut taking place 6 inches from the edge if measured lengthwise. Similarly, the second bit in our representation represents a possible cut taking place 12 inches from the edge if measured widthwise, and so forth. The last 6 bits mean similar things, but they are measured along the width of the solar panel. An individual has this 16 bit array plus an array of height and angles for every sub-plate. The height indicates the position of sub-plate with respect to the ground or beginning of the tree, which is in range of 32 inches to 72 inches. While surface tilt angles and orientation angles indicate angle of the panel with respect to horizontal axis and in which direction the the sub-plate will face respectively. Surface tilt angles are ranged between -90 to 90 and orientation angles are ranged between 0 to 360.

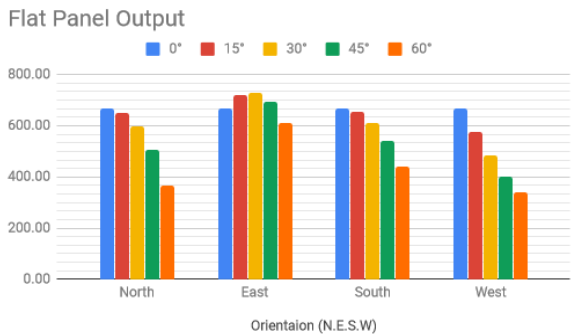

Figure 1. Flat Panel Energy Output in Watts

For the fitness function, we have used pvlib, a python package. It is an open source python library which provides a reliable benchmark implementation for PV system models. The function is capable of dynamically calculating the energy output at different latitudes and longitudes with different surface tilts and orientations. This function assumes output for a regular solar panel of 65 x 39 inches. For each individual, we had to scale the function's output to proportionately match the area of each subplate. The function is also capable of calculating the output at different time periods. We chose the month of August with the time interval from 11:00 AM to 7:00 PM, calculating at every hour for that day. As for the location, we decided to do our experiment in Athens, GA [33.957409, -83.376801]. This function bases it's calculation off of the Sandia Photovoltaic Array Performance Model (SAPM) and the California Energy Commission (CEC) model.

Apart from the fitness function, we made two more helper functions. One of them is bit resolution function and the other one is a conflict deduction function. The bit resolution function ensured that there would not be more than 6 cuts in an individual. By doing this, there will be at most 16 sub-plate. Whereas conflict deduction checks if there are any collisions between the solar sub-plates during the arrangement on tree. For any conflict, an individual's fitness will be decreased by 50 Watts for every conflict. This encourages the evolutionary algorithms to evolve systems which arrange the sub-panels such that they will not overlap. The threshold values for height, surface tilt angle and orientation angle are 20, 90 and 45 respectively.

**A. Genetic Algorithm**

A steady state Genetic Algorithm (GA) was ran for 4000 fitness evaluations such that the average fitness of the population was recorded every 100 fitness evaluations. It had a population size of 100, and 30 independent runs were made.

The GA started with a population of random individuals. The fitness of each individual was calculated and stored. The parents of a child were selected via 2-parent Tournament Selection. Once selected, the parents underwent one-point crossover. To result in two children. Each child was subjected to a mutation, where up to 10 genes could be mutated. The chance that one of those 10 genes would be mutated was pm=0.5. If it bit was selected to be mutated, it was flipped. If a height or angle gene was selected to be mutated, they were given a random value in a range that was appropriate for that kind of parameter. After this, each child underwent a resolution step where the number of ones (cuts) in the individual's was restricted to 6. This keeps the individual from making too many sub-panels. After this, the two children replaced the two worst individuals in the current population. This process was repeated until 4000 fitness evaluations were done.

**B. Evolution Strategy**

The Evolutionary Strategy (ES) had a population of 50 randomly generated individuals with their own sigma value. As the generations continued on, the results of the average individuals were recorded every 800 fitness evaluations. This continued on for 30 different attempts and the best individual, along with their fitness and sigma values were recorded. The ES had 10,000 fitness calculations per the 30 attempts to allow enough time for the population to search as many possible solutions in the search space.

Evolutionary Strategies use different methods for mutation and survivor selection that separate them from the rest of the evolutionary algorithms. Representation and recombination only very slightly than the other methods presented in this paper. Representation adds only one randomly generated sigma value for each individual and recombination takes one of the two parents' positions randomly resulting in one offspring. After recombination, the individual's sigma value is mutated and then used to mutate the individual. Once the individual is mutated, it's fitness is calculated and the best individual and it's sigma is returned. There are two learning parameters that go into mutating the sigma value, $\tau'$ and $\tau$, that are proportionate to $1/(2n)^{1/2}$ and $1/(2n^{1/2})^{1/2}$ respectively. The mutated sigma value, $\sigma_i' = e^{\tau'*N(0,1)+\tau*N_i(0,1)}$ and the newly mutated individual, $x_i' = x_i + \sigma'*N_i(0,1)$.

Lastly, the ES has some useful techniques with the survivor selection, $(\mu,\lambda)$ and $(\mu+\lambda)$. In the comma selection strategy we make $\lambda$ children and select the $\mu$ best of the children, leaving the parents behind in the old generation. The plus strategy, we still create $\lambda$ children but we will choose the best $\mu$ individuals for the next generation from both the parents and the $\lambda$ children. Both strategies have the strengths and weaknesses. The comma can keep up with changing optimums because it leaves all parents behind, good or bad, where the plus will struggle since it can potentially keep some of the bad parents. We chose to run both strategies since this is not a moving optimum problem and wanted to explore the search space with different approaches. So out specific strategies were (50, 350) and (50+350).

**C. Evolutionary Programming**

An Evolutionary Program (EP) was ran for 4000 fitness evaluations, and after every 100 fitness evaluations, the average fitness of the population was recorded. For the sake of time, the population size was restricted to 10. The program was run 30 times, and the best solution found for each attempt was recorded.

The EP started with a population of random individuals. Each individual has a corresponding list of sigma values and a corresponding fitness that was stored. Thus, every gene had a sigma value, and every individual had a fitness associated with it. During every iteration, each individual in the population was a parent and made a child individual, which itself had a sigma list and fitness value associated with it. The child was made by mutating every gene in the parent. For the bits, each bit had a pm=0.2 chance of being flipped. For the rest of the individual, the new value was found with the following equation: newGene = parentGene + (sigma)(N(0,1)) where N(0,1) is the Gaussian function with a mean of 0 and a standard deviation of 1. The child's sigma list was created in a similar fashion. Every new sigma value was created with the equation: newSigma=(parentSigma)(1+(learningRate)(N(0,1)). Note that the minimum value of sigma was restricted to be 0.5. By mutating the sigma, the EP self-adapted to have the best sigma for each gene over time. Note that it is imperative that the sigma list is mutated before the individual is mutated. Also note that crossover was not used.

Out of the parents and children, the $\mu$ best individuals were kept. To find the $\mu$ best, each individual engaged in q=10 competitions with a random individual for each competition. If the individual had a higher fitness, it gained a point. This process was done for each individual. Then the individuals with the $\mu$ highest scores were kept and made into the next population. This process was repeated until 4000 fitness evaluations were done.

## IV. Results

We have achieved comparatively similar results compared to flat solar panel. Out all three of our evolutionary algorithms (Genetic Algorithm, Evolutionary Programming and Evolution Strategy) Genetic Algorithm outperforms the rest. As shown in table 2 and figure 2 the average best and global best for genetic algorithm is 621.36 W and 623.25 W respectively. The evolutionary strategies, plus and comma, took second place with the global best of 592.59 and 593.58. Their average best being 507.92 and 505.989. Lastly, our experiment yielded a global best of 591.63 and average best of 555.05 in last place with the evolutionary programming. Figure 2 shows the average fitness each method found.

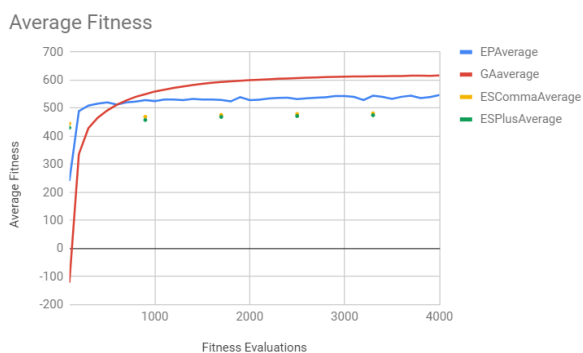

Figure 2: The average fitness of each method. This graph was made by averaging the results of the thirty runs. Note that the ESComma and ESPlus methods were not given a line so they would be easier to see.

To make the comparison with flat solar panel more reliable, we have computed the output of flat solar panel with same pvlib fitness function as we did in evolutionary algorithms furthermore the location and the time were also same. Table 1 and Figure 1 shows the output of the flat solar panel with respect to different surface tilt and different orientations. The highest energy output flat solar panel got was 728.71 W at 30° surface tilt while facing east. Location and time for testing the flat solar panel and the solar tree were same.

TABLE 2: The best fitness found for each method in Watts.

|  | GA | EP | ES |
|---|---|---|---|
| **Average Best Fitness Over 30 attempts** | 621.36 | 555.05 | $(\mu, \lambda)$ 505.989 $(\mu + \lambda)$ 507.92 |
| **Global Best Fitness over 30 attempts** | 623.25 | 591.63 | $(\mu, \lambda)$ 593.58 $(\mu + \lambda)$ 592.59 |

As shown in figure 3 and 4, we have built a 3D model of our best individual of steady state Genetic Algorithm. The model was built in TinkerCAD, an open source AutoCAD software. Figure 3 is the top view of the solar tree and figure 4 is front view of the solar tree.

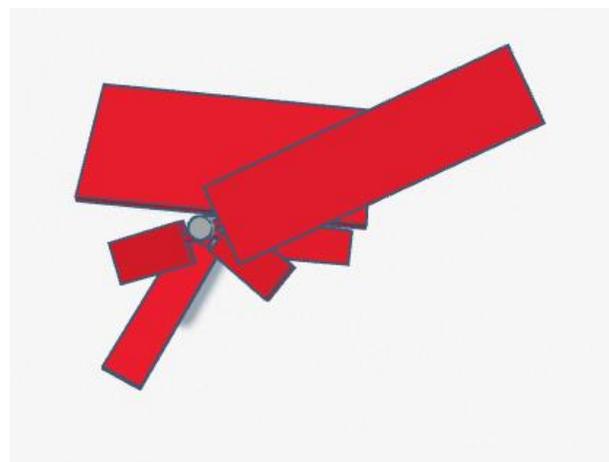

Figure 3. Top view of Solar Tree

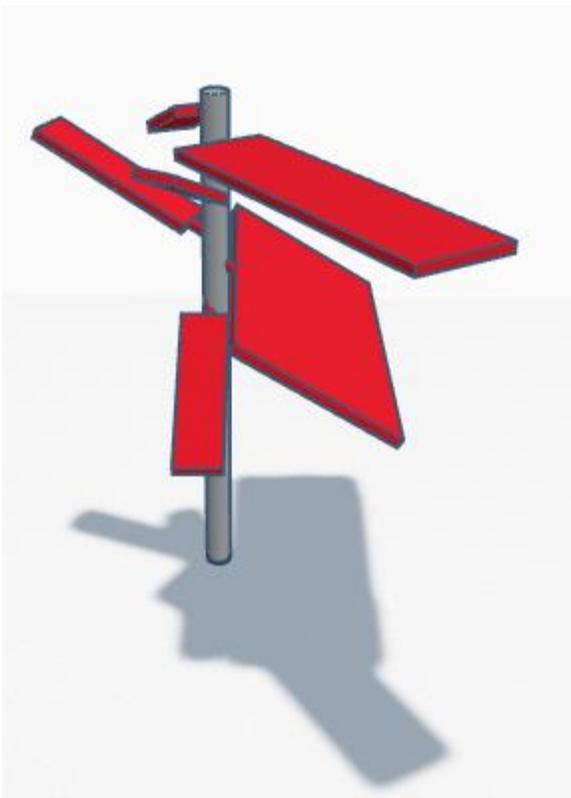

Figure 4. Front view of Solar Tree

TABLE 3. Statistical T - test (2 - tailed)

| T-test (2-tailed) | p-values |
|---|---|
| GA and EP | 0.00849 |
| ES (comma) and (plus) | 0.00385 |

Table 3 shows the reliability of our evolutionary algorithms. We have done T - test to show the statistical significance of comparisons on all three evolutionary algorithms. The results of the t-test align with our experiment. The GA performed significantly better than the EP and similarly for the test for our ES. We could not compare the ES against the GA and EP accurately because the ES ran for a shorter amount of generations each attempt that the other evolutionary algorithms. It would not be an accurate conclusion due to the imbalance of data.

## V. Conclusion

In this paper we tried to optimize the solar panel design in a tree shape in order to get the maximum amount of solar irradiance. We have applied a genetic algorithm, evolution strategy and an evolutionary programming on the same problem to compare which evolutionary algorithm is most suitable for the solar tree. As shown in results genetic algorithm outperforms the other evolutionary algorithm. Although our EAs got comparatively similar results compared to flat solar panels, our solar tree could not beat the flat solar panel. Some of the key points which might be the reason for getting less fitness. One of them being the penalty deduction. Our conflict deduction function deduce 50 fitness from an individual, if there is a conflict between two sub-plates in solar tree. Furthermore we didn't take the reflection of sunlight into consideration, this will affect the fitness itself.

Future work would be to include a better sophisticated penalty function to account for the overlap of the sub-plates. We only tested our panels here in Athens, so to try different places across the globe to see if this approach can outperform flat solar panels in locations where they tend to not do so well. We also assumed the default weather conditions that the pvlib function had, thus testing how the solar tree performs in other environments could change the output of the solar tree. Lastly, it might be good to see how many of our solar trees can be arranged in the same area as flat solar panels to test the total watt output.

# VI. References

[1] McAdams, D. A., & Li, W. (2014). A Novel Method to Design and Optimize Flat-Foldable Origami Structures Through a Genetic Algorithm. *Journal of Computing and Information Science in Engineering*, *14*(3), 031008.

[2] Kowaliw, T., Dorin, A., & McCormack, J. (2012). Promoting creative design in interactive evolutionary computation. *IEEE transactions on evolutionary computation*, *16*(4), 523.

[3] Koutroulis, E., Kolokotsa, D., Potirakis, A., & Kalaitzakis, K. (2006). Methodology for optimal sizing of stand-alone photovoltaic/wind-generator systems using genetic algorithms. *Solar energy*, *80*(9), 1072-1088.

[4] Peraza-Hernandez, E. A., Hartl, D. J., Malak Jr, R. J., & Lagoudas, D. C. (2014). Origami-inspired active structures: a synthesis and review. *Smart Materials and Structures*, *23*(9), 094001.

[5] Hansen, Clifford & Klise, Katherine & Stein, Joshua & Ueda, Yuzuru & Hakuta, Keiichiro. (2014). Photovoltaic System Model Calibration Using Monitored System Data.

[6] Einstein, A. (1905). Indeed, it seems to me that the observations regarding" black-body radiation," photoluminescence, production of cathode rays by ultraviolet. *Annalen der Physik*, *17*, 132-148.

[7] http://www.nbcnews.com/id/35593146/ns/technology_and_science-innovation/t/origami-boosts-solar-panel-productivity/#.XAw_7hBRf_j

[8] Benghanem, M. (2011). Optimization of tilt angle for solar panel: Case study for Madinah, Saudi Arabia. *Applied Energy*, *88*(4), 1427-1433.